\documentclass[sigconf]{acmart}
\AtBeginDocument{%
  }

\setcopyright{acmlicensed}
\copyrightyear{2026}
\acmYear{2026}
\acmDOI{XXXXXXX.XXXXXXX}
\acmConference[SIGKDD'26]{32nd ACM SIGKDD Conference on Knowledge Discovery and Data Mining}{August 9--13,
  2026}{Jeju, Korea}
\acmISBN{978-1-4503-XXXX-X/2018/06}

\usepackage{tcolorbox}
\usepackage{enumitem}
\usepackage{subcaption}
\usepackage{booktabs}   
\usepackage{pifont}     
\usepackage{graphicx}   
\usepackage{array}      
\usepackage{tabularx}    
\usepackage{diagbox}
\usepackage{adjustbox}

\newcommand{\cmark}{\ding{51}} 
\newcommand{\rot}[1]{\rotatebox{70}{\small\itshape #1}}

\begin{document}

\title{Riemannian Geometry Speaks Louder Than Words: 
From\\ Graph Foundation Model to Next-Generation Graph Intelligence}


\author{Philip S. Yu}
\affiliation{%
  \institution{University of Illinois Chicago}
  \city{Chicago}
  \country{USA}}
\email{psyu@uic.edu}

\author{Li Sun}
\affiliation{%
  \institution{Beijing University of Posts and Telecommunications}
  \city{Beijing}
  \country{China}}
\email{lsun@bupt.edu.cn}







\renewcommand{\shortauthors}{Yu et al.}

\renewcommand{\shorttitle}{Riemannian Geometry Speaks Louder Than Words}

\begin{abstract}
Graphs provide a natural description of the complex relationships among objects, and play a pivotal role in communications, transportation, social computing, the life sciences, etc. Currently, there is strong agreement that Graph Foundation Models (GFMs) are essential for advancing graph learning, yet considerable disagreement persists on how to build a powerful, general-purpose GFM analogous to Large Language Models (LLMs). Graph Neural Networks (GNNs) exhibit limitations in memory retention and principled interpretability when confronted with multi-domain pretraining and adaptation. The challenge  of graph serialization hinders the direct application of LLMs, as the words struggle to capture the structural complexity and diversity inherent in graphs. In contrast, Riemannian geometry offers an elegant mathematical framework for modeling structures, while remaining compatible with graph semantic learning, even with LLMs. In this paper, we argue that, for graphs, \textbf{Riemannian geometry speaks louder than words}, and lay out the foundational principles for GFM. Reimagining with Riemannian geometry, we introduce a blue sky idea--\textbf{Riemannian Foundation Model} (RFM)--that opens a new pathway for capturing complex structural patterns and uncovering cross-domain generalities. RFM emphasizes intrinsic graph geometry and embodies endogenous capacities for structural inference and generation, moving beyond mere representation-space switching. Accordingly, we outline a progressive agenda that begins with universal structural understanding through intrinsic geometry, and then rebuilds LLM with a Riemannian engine for general-purpose graph modeling and beyond. Thus, RFM enables a paradigm shift from designing graph models to solving graph-structured applications with RFM agents, unlocking the next-generation graph intelligence. 
\end{abstract}





\maketitle


\begin{figure}[H]
\vspace{-0.2in}
\centering
\begin{minipage}{0.55\linewidth}
    \centering
    \includegraphics[width=\linewidth,page=1]{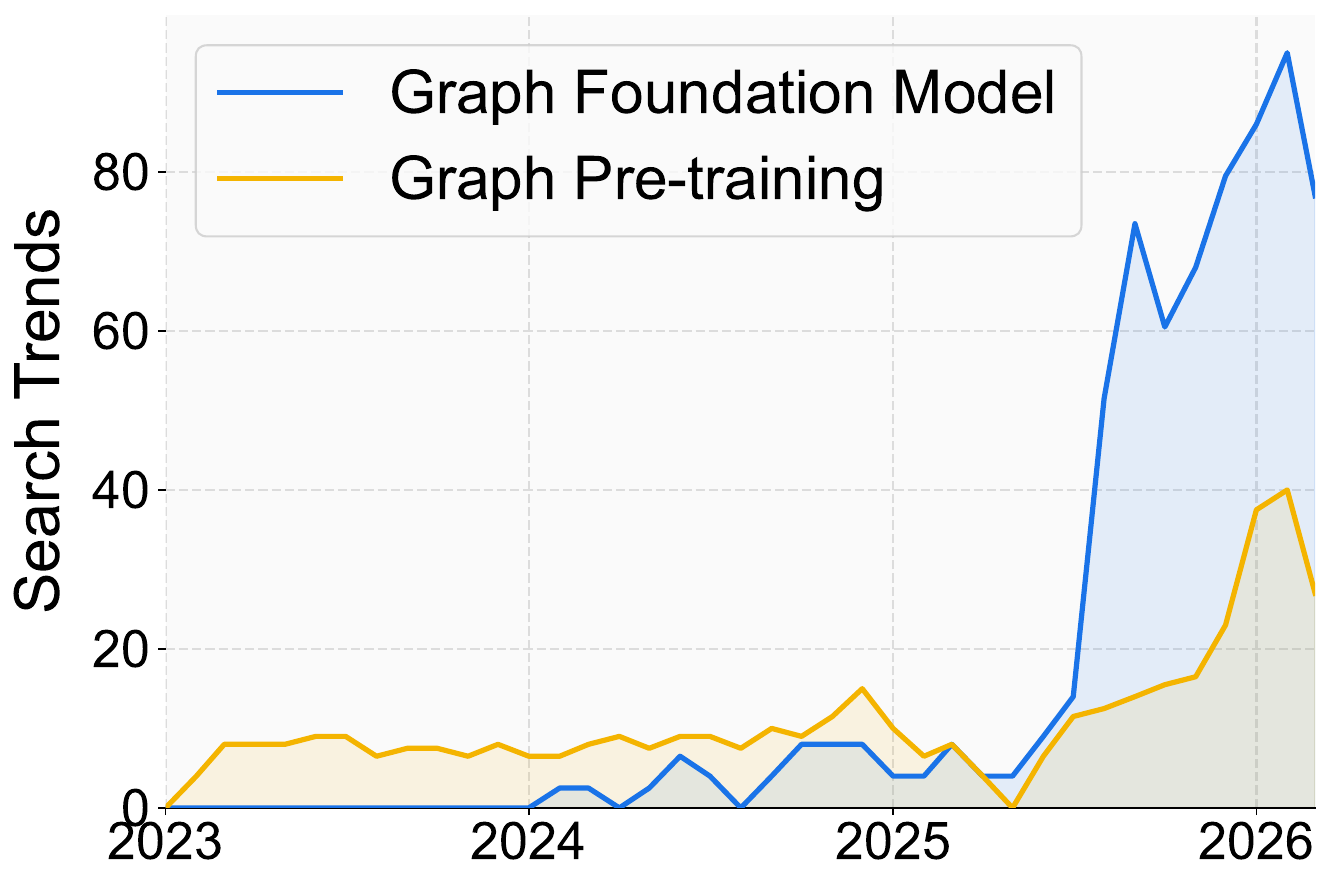}
\end{minipage}
\hfill
\begin{minipage}{0.43\linewidth}
    \centering
    \includegraphics[width=\linewidth,page=1]{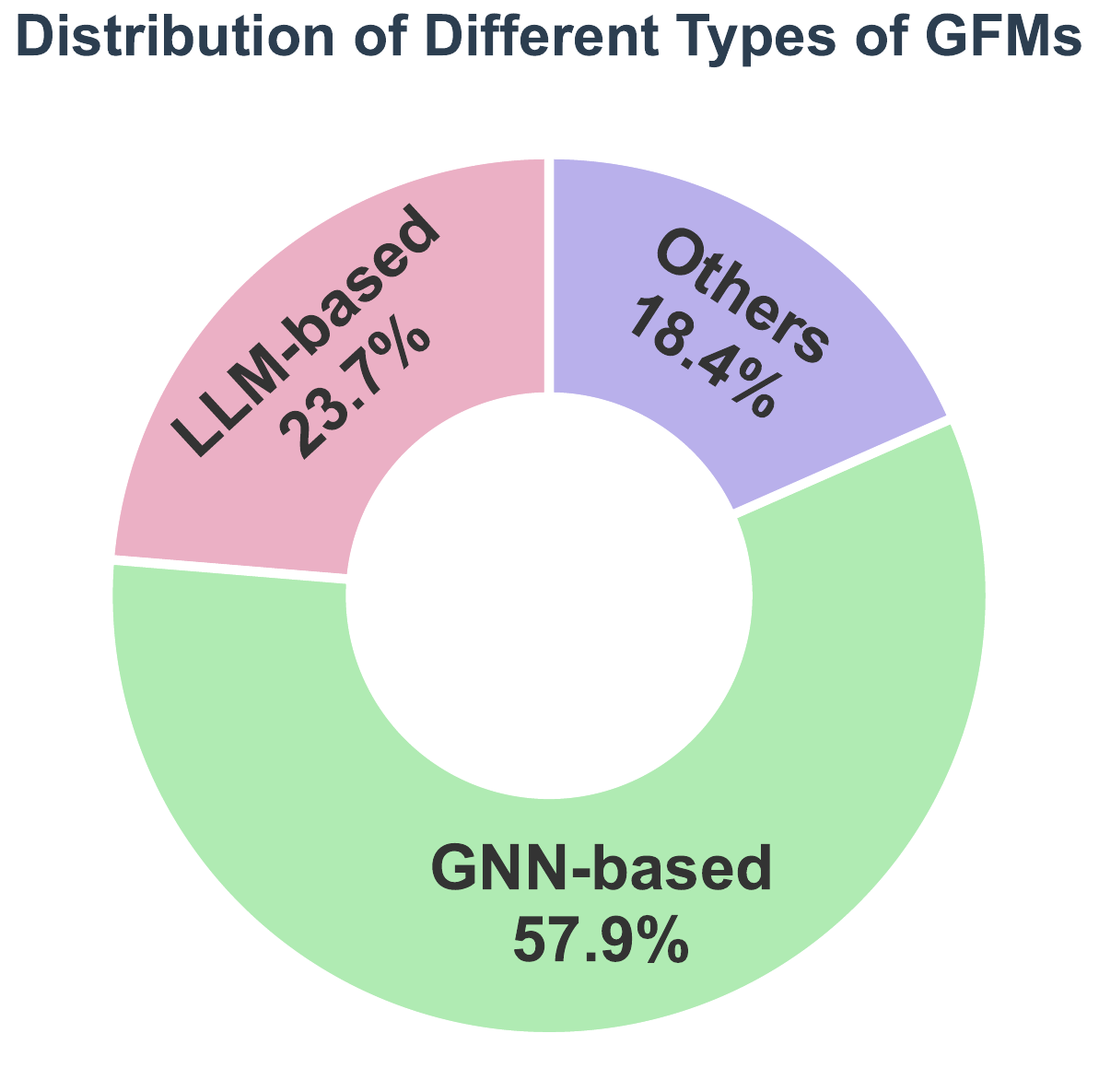}
\end{minipage}
\vspace{-0.05in}
\caption{Left: Google Trends for Graph Foundation Model and Graph Pre-training. Right: Proportion of existing graph foundation model papers by architecture type.}
\label{trand}
\end{figure}

\begin{table*}[t] 
\centering 
\caption{Capability Matrix of Graph Foundation Models (GFMs) across Diverse Tasks} 
   \vspace{-0.1in}
\label{tab:gfm_capability_matrix} 
\small 
\renewcommand{\arraystretch}{1.3} 
\setlength{\tabcolsep}{3pt} 
\begin{tabularx}{\textwidth}{>{\small\itshape\raggedright\arraybackslash}m{2.8cm}| *{20}{>{\centering\arraybackslash}X}} 
\toprule 
& \rot{OpenGraph~\cite{xia2024opengraph}} & \rot{OFA~\cite{liu2024one}} & \rot{GraphAny~\cite{zhao2025fully}} & \rot{RAGraph~\cite{jiang2024ragraph}} & \rot{GCOPE~\cite{zhao2024all}} & \rot{GRAVER~\cite{yuan2025graver}} & \rot{GIT~\cite{wang2025towards}} & \rot{RiemannGFM~\cite{sun2025riemanngfm}} & \rot{BRIDGE~\cite{yuan2025much}} & \rot{GraphGPT~\cite{tang2024graphgpt}} & \rot{AnyGraph~\cite{xia2024anygraph}} & \rot{AutoGFM~\cite{chen2025autogfm}} & \rot{SA$^2$GFM~\cite{shi2025sa}} & \rot{MDGPT~\cite{yu2024text}} & \rot{SAMGPT~\cite{yu2025samgpt}} & \rot{FedGFM+~\cite{zhu2025towards}} & \rot{FedBook~\cite{wu2025fedbook}} & \rot{UniGraph~\cite{he2024unigraph}} & \rot{MDGCL~\cite{zhao2025towards}} & \rot{GFT~\cite{wang2024gft}} \\ 
\midrule 
Node Classification & \cmark & \cmark & \cmark & \cmark & \cmark 
& \cmark & \cmark & \cmark & \cmark & \cmark & \cmark & \cmark & \cmark & \cmark & \cmark & \cmark & \cmark & \cmark & \cmark & \cmark \\ 
Link Prediction     & \cmark & \cmark & & & & & \cmark & \cmark & & \cmark & \cmark & \cmark & & & & \cmark & \cmark & \cmark & & \cmark \\ 
Graph-level Task    & \cmark & \cmark & \cmark & \cmark & & \cmark & \cmark & \cmark & \cmark & \cmark & & \cmark & \cmark & \cmark & \cmark & \cmark & \cmark & \cmark & \cmark & \cmark \\ 
Graph Generation    & & & & & & & & & & & & & & & & & & & & \\ 
\bottomrule 
\end{tabularx} 
\end{table*}

\vspace{-0.1in}
\section{Introduction}

Graphs are ubiquitous, and graph learning is pivotal to the advancement of data mining and artificial intelligence. The emergence of Large Language Models has catalyzed a profound paradigm shift in this domain, driving the field from specialized, fragmented models toward universal foundation models. Recent efforts to replicate the success of foundation models in graph learning have ignited growing interest in Graph Foundation Models —a class of expressive, general-purpose, pretrainable neural architectures for graphs, as illustrated in Figure \ref{trand}.

Unlike texts and images, the non-Euclidean nature of graphs charts a distinct developmental course.
The road to GFM is currently divided as in Figure \ref{trand}, with both prevailing approaches facing inherent challenges.
One line of work extends Graph Neural Networks via multi-domain pretraining \cite{xu2024graphfm,lachi_graphfm_2024}.
However, these models are fundamentally constrained by local message passing and fixed inductive biases, such as Euclidean representation spaces, failing to capture structural diversity.
Their limitation in memory retention also impedes the discovery of cross-domain generalities.
Another line of work leverages LLMs as the cornerstone, demonstrating initial success primarily on text-attributed graphs \cite{liu2024one}.
This paradigm focuses on devising word-like graph serialization techniques to accommodate LLMs, yet it inherently struggles to traverse the vast structural diversity inherent in graphs \cite{yuan2025graver,zhang2023graph}.
Also, representing graphs  as ``word'' sequences poses a fundamental challenge in capturing their intricate structural complexity \cite{zhou2025each}.

We argue that, \textbf{for graphs, Riemannian geometry speaks louder than words.}
To date, neither LLM-based nor GNN-based approaches have identified the graph generalities or invariances that underpin universal graph modeling.
Standing at a crossroads, we return to the essence of graphs—the discrete analogs of manifolds—and thereby illuminate a new direction: a Riemannian geometry approach.
Indeed, Riemannian geometry provides a rigorous foundation for graph modeling. In this elegant framework, there exists a powerful toolkit to build GFMs, including: 1) a rich family of manifolds to capture structural diversity, 2) the concept of curvature to delve into structural complexity, 3) the moving frame method \cite{sunmulti} to transfer knowledge across domains, and 4) the construction of vector bundles \cite{tu2011manifolds} to integrate graph structure and semantics, creating a synergistic interface to integration with LLMs.

Grounded in Riemannian geometry, we put forward a blue—sky idea of \textbf{Riemannian Foundation Model}, aimed at general-purpose graph learning and beyond. While a significant body of work on non-Euclidean graph learning exists, it typically focuses on selecting extrinsic representation spaces, such as hyperbolic spaces, tailored to specific graphs \cite{sun2021hyperbolic,sun2024lsenet,wang2020hyperbolic}. Moving beyond the mere switching of representation spaces, RFM emphasizes graph intrinsic geometry such as flatness, symmetry, connectivity, and anisotropy. It offers a principled framework to uncover graph generalities, interpret knowledge transfer across domains, and endow models with endogenous capacities for structural inference and generation. To realize this vision, we outline a progressive agenda toward RFM, thereby opening the next-generation graph intelligence. Analogous to agents for LLMs, RFM agents are envisioned as the key to solving graph-structured applications, from code generation to drug design.

\begin{figure*}[t]
\centering
\begin{subfigure}{0.48\linewidth}
    \centering
    \includegraphics[width=\linewidth,page=1]{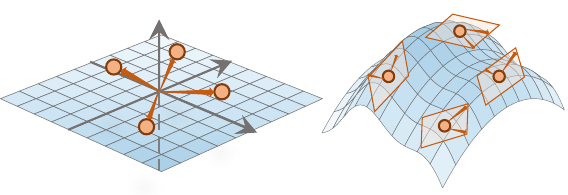}
    \caption{vector bundle}
    \label{fig:vector bundle}
\end{subfigure}
\hfill
\begin{subfigure}{0.48\linewidth}
    \centering
    \includegraphics[width=\linewidth,page=1]{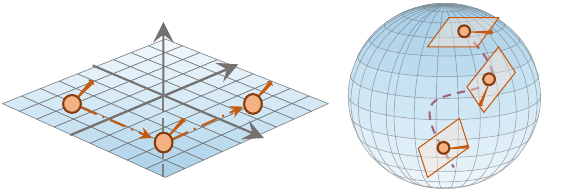}
    \caption{Parallel Transport}
    \label{fig: Parallel Transport}
\end{subfigure}
    \vspace{-0.1in}
\caption{Illustration of parallel transport and vector bundle.}
    \vspace{-0.1in}
\end{figure*}

\vspace{-0.2in}
\section{Current Challenges}

\subsection{Are GNNs Generalizable?}

Graph Neural Networks, powered by the message-passing mechanism, have become the dominant solution for graph representation learning~\cite{kipf2016semi,hamilton2017inductive,velivckovic2018graph,chen2020simple}.
Recently, a surge of research has sought to extend GNNs into the realm of foundation models~\cite{xia2024opengraph,liu2024one,zhao2024all,liu2024one}, yet a fundamental mismatch persists between the two.
While GNNs are typically trained on specific graphs by task-specific objectives, Graph Foundation Models aspire to achieve general-purpose modeling across arbitrary graphs through large-scale pretraining. With the advent of self-supervised techniques such as graph contrastive learning \cite{bevilacqua2025holographic,wang2025towards,hou2024graphalign}, GNNs have reduced their dependence on target tasks by learning from the data themselves. 
Nevertheless, GNNs still struggle with structural shifts in data and suffer from inherent scalability issues, e.g., the well-known oversmoothing issue preventing the network from going deep \cite{keriven2022not}.

We posit that the core challenge lies in accommodating the vast diversity of the graph domain. Graphs exhibit wildly different topological patterns, ranging from the rigid bonds in molecules to the chaotic links in social networks \cite{klaser2024minimol,sypetkowski2025scalability,zhang2024unimot}, yet conventional GNNs rely on a single, fixed Euclidean representation space. This severely limits their expressiveness in differentiating structural patterns. The problem is exacerbated during cross-domain generalization. While multi-domain pretraining has been proposed to enhance generalization ~\cite{he2025generalizing,yu2025samgpt,wang2025multi,yuan2025much}, catastrophic forgetting remains problematic. Notably, jointly capturing shared knowledge while preserving distinct structural semantics remains inherently challenging \cite{sun2025riemanngfm}, and a principled interpretive framework is still lacking. 

We argue that, \textbf{although multi-domain pretraining marks progress, the GNN mechanism itself reflects a deeper inability to represent diverse topologies and the structural semantics they encode.}

\vspace{-0.1in}
\subsection{Can LLMs Capture Graph Geometry?}

LLMs have revolutionized textual understanding and generation, prompting a paradigm shift toward   developing LLM-based GFMs. A central question arises: can linguistic tokens—``words''—adequately capture the structural complexity inherent in graphs?  Within this line of research, much effort has focused on graph serialization to ensure compatibility with LLM architectures. Existing serialization methods include graph random walks~\cite{ye2024language,hu2023beyond,sun2025graphicl,tang2024graphgpt,zhang2024graphtranslator}, path sampling~\cite{tang2025toward,kim2025flock}, and Hamiltonian cycles rooted in topological theory \cite{bosnic2023finding}. These methods aim to equip LLMs with structural knowledge by exposing them to local graph contexts. However, a traversal offers only a fragmented, path-dependent view of the graph, while falls short of capturing the high-order structural patterns that govern the entire topology. We argue that \textbf{universal graph modeling via LLMs remains a distant goal, hindered by the intractability of traversing diverse graph domains and the limitation of representing structural semantics through words alone.}

Mimicking the advances in LLMs, retrieval-augmented frameworks have also been proposed for GFMs~\cite{li2025large,jiang2024ragraph,wang2025rag4gfm}.
These methods typically retrieve subgraphs based on embeddings, conflating semantic proximity with topological relevance. In practice, nodes that are structurally interdependent may share little semantic resemblance, while semantically similar nodes are often topologically distant. Consequently, preserving structural dependencies becomes a critical bottleneck, yet it remains essential for achieving  universal graph modeling. \textbf{Furthermore, the ability for graph generation remains conspicuously absent among state-of-the-art GFMs (as shown in Table \ref{tab:gfm_capability_matrix}), despite generative modeling ability being a hallmark of LLMs.}
 \vspace{-0.05in}
\section{Riemannian Geometry Speaks Louder Than Words for Graphs}

\vspace{0.05in}
\subsection{Rethinking GFM}

GFMs represent a class of expressive, general-purpose, and pretrainable neural architectures designed for graphs.
At the heart of any foundation model lies the ability to identify generalities (or invariances) which enable broad adaptability.
The success of LLMs is rooted in a shared word vocabulary that spans across texts.
However, words alone fall short of capturing the structural complexity of graphs, as discussed above.
The core challenge in the graph domain is thus to uncover the generalities that govern universal graph modeling and to establish a principled framework for understanding how structural semantics transfer across different graphs.
\textbf{Standing at a crossroads, we peopose to open a new direction for building GFMs.}
Before unfolding this blue-sky idea, we first return to the essence of graphs.

\vspace{-0.05in}
\subsection{New Geometric Toolkits}

Graphs can be viewed as discrete analogs of manifolds, and Riemannian geometry provides a rigorous foundation for modeling them.
This geometric framework naturally equips us with a powerful toolkit for GFMs. We therefore argue that, \textbf{when it comes to graphs, Riemannian geometry speaks louder than words.}

\subsubsection*{\textbf{Manifolds and Vector Bundle.}} 
A manifold is a space that locally resembles flat space but may curve globally, providing a natural geometric foundation for GFMs to model the intrinsic structure of graphs beyond fixed coordinate assumptions.
As shown in Figure \ref{fig:vector bundle}, a vector bundle assigns an independent vector space to each point on the manifold, allowing representations to reside in domain-specific vector spaces. 
This design enables flexible and expressive representation learning while preserving structural integrity, making it well-suited for multi-domain graph modeling.

\subsubsection*{\textbf{Riemannian Metric and \'{E}lie J. Cartan's Moving Frame Method}} 

A Riemannian metric equips a manifold with a smoothly varying inner product that defines local distances and angles, allowing the geometry to accommodate domain-specific structural patterns without distorting intrinsic graph structure.
A moving frame provides a local coordinate basis that evolves along the manifold and aligns representations by adapting to the local metric.
Together, the metric and the moving frame allow the geometry to adapt across graph domains while maintaining consistent representations.

\subsubsection*{\textbf{Curvature, Levi-Civita Connection and Parallel Transports.}} 

A connection defines how vectors at different points on a manifold are compared. 
As shown in Figure \ref{fig: Parallel Transport}, parallel transport moves vectors between points while preserving geometric relationships, providing a principled geometric basis for message passing by transporting node representations into comparable spaces prior to aggregation.
Furthermore, curvature measures the degree to which a manifold deviates from flatness. 
When representations are transported along different paths connecting the same nodes, the resulting curvature may differ, indicating geometric inconsistency in the underlying space.
In this sense, curvature acts as an intrinsic geometric indicator of coherent inference.

\begin{figure*}[htbp]
    \centering
    \includegraphics[width=1.0\textwidth]{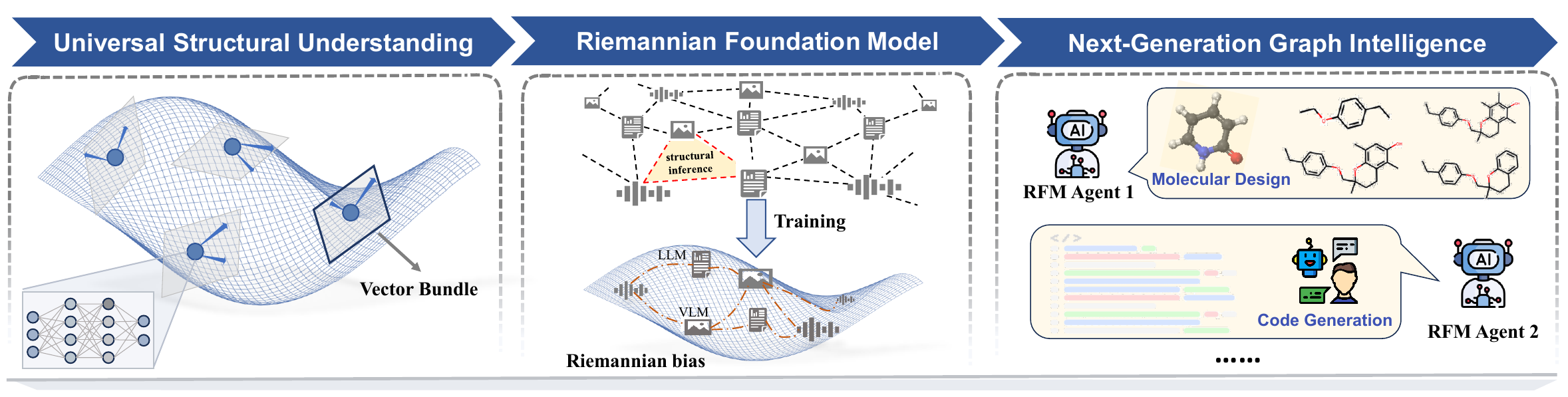}
    \vspace{-0.2in}
    \caption{An illustration of the road to next-generation graph intelligence.}
    \label{fig:framework}
        \vspace{-0.1in}
\end{figure*}

\vspace{-0.1in}
\subsection{Blue Sky Idea }

\subsubsection*{\textbf{Our Recommendation.}} 

As introduced above, Riemannian geometry provides a rigorous foundation for graph modeling, illuminating a new direction for building GFM, a Riemannian geometry approach. In this paper, we put forward a blue sky idea of \textbf{Riemannian Foundation Model (RFM)}. This perspective shifts the focus of graph learning from designing representations to discovering geometry. 
Moving beyond the mere switching of representation spaces, RFM emphasizes graph intrinsic geometry such as flatness, symmetry, connectivity, and anisotropy. The key is to recover the manifold underlying graphs so that generality and structural invariance emerge naturally from the geometry itself, rather than being imposed through handcrafted architectures. This broader geometric perspective allows models to capture structural patterns that vary across domains within a unified mathematical framework. It offers a principled framework to uncover graph generalities, interpret knowledge transfer across domains, and endow models with endogenous capacities for structural inference and generation.

\vspace{-0.05in}
\subsubsection*{\textbf{Connection to Related Approaches.}} 

We discuss the inherent difference among RFM, geometric deep learning and non-Euclidean graph learning.
\textbf{Geometric deep learning} \cite{bronstein2017geometric} focuses on explicit geometric data such as point clouds or meshes, modeling distances, angles, and 3D coordinates in Euclidean space. 
 \textbf{Non-Euclidean graph learning} extends graph neural networks to curved spaces, such as hyperbolic or spherical manifolds, and has proven effective in modeling hierarchical and complex relational structures \cite{liu2019hyperbolic, do1992riemannian}.
This line of work centers on representation learning, and recent efforts introduce advanced manifold constructions tailored to specialized graphs \cite{deng2023fmgnn,sun2025riemanngfm,li2025multi}.
In contrast, \textbf{RFM} emphasizes the intrinsic properties of the manifold itself such as flatness, symmetry and connectivity, independent of any embedding. 
Rather than operating in fixed coordinate spaces, RFM aims to uncover the underlying generalities and geometric structure governing real-world graphs across multiple domains.


\vspace{-0.1in}
\section{A Progressive Agenda}

Grounded in Riemannian geometry, we outline a progressive agenda towards universal  graph modeling and beyond, as sketched in Figure    \ref{fig:framework}.
Specifically, we begin with universal structural understanding through intrinsic geometry, and then rebuild LLM with a Riemannian engine featured by the endogenous capability of structural inference, unlocking the next-generation graph intelligence.

\vspace{-0.1in}
\subsection{Universal Structural Understanding}
Here, we active a previously ignored concept of intrinsic geometry for pretraining and adaptation across the graph domains.

\vspace{-0.05in}
\subsubsection*{\textbf{Intrinsic Geometry \& Invariant Structures.}}
We first address a fundamental objective in universal graph modeling: identifying generalizable patterns across graph domains. While extensive research has explored potential candidates for such generalities, including motifs \cite{sun2025riemanngfm}, tree message-passing schemes \cite{wang2024gft,wang2025towards}, graphons \cite{yuan2025graver}, and structural vocabularies \cite{bo2025quantizing,jiang2024ragraph}, these approaches often struggle with adaptability. Specifically, they lack a mechanism to determine which pretrained structural patterns remain applicable to specific target graphs. This often results in inconsistent functional behavior and suboptimal performance across different contexts. In contrast, Riemannian geometry provides the framework of vector bundles to capture intrinsic geometry, effectively uncovering structural invariants that persist across disparate domains.

\vspace{-0.05in}
\subsubsection*{\textbf{Princepled Knowledge Transfer.}} 

Second, we establish a principled framework to understand how structural semantics transfer across different graphs. 
Rather than relying on fixed geometries like hyperbolic or spherical spaces, RFM learns the smooth, underlying manifold directly from node representations to ensure a well-defined Levi-Civita connection.
Given such a manifold, the concept of parallel transport can be leveraged to understand how knowledge is integrated or transferred across domains. Within this vector bundle construction, the derivation of invariant structures offers a measure of transfer effort and specifies the the mechanism by which behaviors learned from substructures during pretraining become applicable to unseen graphs.

\vspace{-0.1in}
\subsection{Riemannian Foundation Model}

We outline the blueprint of RFM for the future graph intelligence.

\vspace{-0.05in}
\subsubsection*{\textbf{Integrating LLMs.}} 
We propose a unified path for integrating GFMs and LLMs into a cohesive whole. While universal graph modeling demands a deep understanding of both structural patterns and attribute semantics, these components exhibit distinct geometric properties: graph structures are inherently Riemannian, whereas attributes like text and images typically reside in Euclidean spaces.
Simultaneously modeling such heterogeneous modalities remains a non-trivial task.
To address this, we leverage the concept of vector bundles in Riemannian geometry. This principled framework reconciles diverse geometries by allowing LLMs to operate effectively within Euclidean fibers. Such a construction demonstrates that vector bundles, while preserving structural invariance, are also fundamentally compatible with the integration of LLMs.

\vspace{-0.05in}
\subsubsection*{\textbf{Endogenous Structural Inference.}} 
We introduce endogenous structural inference as a hallmark of RFM. 
In our construction, large-scale post-training is conducted on a multimodal graph in the real world. By inherently anchoring linguistic tokens to a topological manifold, the generative process is thus aware of complex relationships among objects, equipped with the endogenous capability of structural inference, understanding the rationale underlying the linkages.
While existing GFMs prioritize task-independent representations to achieve general-purpose utility, they often lack insight into the fundamental mechanisms of structure formation.
By contrast, endogenous structural inference empowers RFM with enhanced ability in complex logic and structural reasoning.

\vspace{-0.05in}
\subsubsection*{\textbf{Graph Generation.}}
Here, we discuss graph generation within the framework of RFM, which remains conspicuously absent among current GFMs yet stands as a hallmark of LLMs. Graph generation is vital in a wide range of real-world applications ranging from drug discovery to circuit design. As claimed in the universality, the ability forGraph generation should be included in GFMs, yet it is evidently a distant goal. RFMs inherit the generative modeling ability during the integration of LLMs, and endogenous structural inference enables the model to recover the linguistic tokens as graphs. It not only achieves graph generation but also provides the interpretative clues for such generation.




\begin{tcolorbox}[boxsep=0mm,left=2.5mm,right=2.5mm]
\textbf{Our Message.} 
\textit{The geometric toolkit provided by Riemannian geometry is ready to build GFMs and redefine what a foundation model for graphs can be. RFM identifies the generalities using intrinsic geometry, provides principled interpretation of knowledge transfer, embodies the endogenous capability of structural inference and enables the graph generation which remains conspicuously absent among GFMs. Also, we suggest a wider recognition of \textbf{intrinsic geometry} that governs the structural patterns in the graph domain, instead of the coordinates in the representation space, and emphasize that the \textbf{endogenous  structural inference} should be a key to universal graph modeling and beyond.}
\end{tcolorbox}

\subsection{Graph Intelligence with RFM}

Benefit from RFM, the evolution of AI demands a fundamental shift from linguistic pattern matching to intrinsic structural cognition, no longer viewing the world as mere isolated embeddings.

\vspace{-0.05in}
\subsubsection*{\textbf{Multimodal World Cognition beyond Graphs.}} 

In our envisioned paradigm, the RFM serves as the fundamental medium across diverse modalities, driving next-generation AI from passive perception toward interactive spatial intelligence. As highlighted by recent World models \cite{ha2018world,ding2025understanding}, true world cognition requires understanding not just what entities are, but how they physically and logically interact within a space.
To achieve this, the RFM constructs a universal coordinate system from the environment's underlying relational skeleton. Rather than loosely associating text, vision, and physical states through statistical similarity in an embedding space, the RFM anchors all disparate modalities directly onto this topological skeleton. 
Whether parsing the 3D spatial configuration of a robotic workspace or the logical prediction, the model enforces strict relational constraints. This structural grounding ensures that AI perception fundamentally aligns with human cognition.

\vspace{-0.05in}
\subsubsection*{\textbf{RFM Agents for Graph-structured Applications.}} 

The application paradigm of graph learning will undergo a radical transition mirroring the evolution of LLMs: shifting from training isolated predictive models to deploying general-purpose autonomous agents~\cite{yao2022react}. 
In the next generation, the RFM functions as a central cognitive engine that interacts with the network environment and various soft tools. By perceiving the graph as an actionable space, the structural agent can navigate complex topologies, execute multi-step reasoning, and dynamically invoke structural tools to solve open-ended problems. For instance, in molecular discovery, rather than merely predicting a compound's toxicity, an RFM agent can autonomously traverse the molecular graph to pinpoint toxic subgraphs and orchestrate chemical editing tools to swap functional groups. This transforms graph AI from a static analytical calculator into an interactive and decision-making entity capable of actively optimizing the structures it perceives.

\section{Alternative Views}
This section presents divergent perspective surrounding Graph Foundation Models. 
Current discussions primarily revolve around two central questions: the optimal architecture for graph intelligence, and the nature of its symbiosis with Large Language Models.

\subsection{Architecture of Graph Foundation Model}

Beyond message-passing GNNs, graph transformers have emerged as another prominent architecture for graph modeling, integrating structural inductive biases into standard self-attention mechanisms \cite{he2025generalizing,tang2024higpt,rampasek2022recipe}. 
Recent efforts strive for universal graph modeling through the transformer architecture, exemplified by models such as GDL4LLM \cite{zhou2025each} and GraphFM \cite{lachi_graphfm_2024}.
These approaches conceptualize graphs as a new structural language, discretizing topological data into token sequences compatible with transformer backbones.
Theoretically, such models can learn to simulate graph connectivity via massive-scale pre-training, suggesting that structural intelligence may emerge from compressing multi-domain features into a shared latent space.
Meanwhile, other researchers attempt to combine text encoding with graph aggregation by nesting GNN components within transformer blocks \cite{tang2024graphgpt,chen2024llaga,liu2024can}.
From this perspective, the success of GFMs hinges on sampling techniques and the efficiency of structural tokenization. However, we contend that this paradigm introduces a fundamental representational bottleneck: by forcing non-Euclidean graphs into a one-dimensional format, these models inevitably fragment global topology. This suggests that achieving a powerful GFM remains a distant goal and underscores the necessity of studying graph structures from a more native geometric perspective rather than treating them as sequences.


\subsection{Collaboration with Large Language Model}

Another debate centers on whether GFM should function as an external plugin for language models or serve as the intrinsic cognitive foundation for all reasoning. 
The prevailing paradigm maintains that LLMs should remain the primary engine of cognition.
In this line, models either translate graph data into text-based prompts (e.g., GraphTranslator \cite{zhang2024graphtranslator}) or project graph embeddings into the language semantic space (e.g., LLaGA \cite{chen2024llaga}). While this offers flexibility, it relegates structural dependencies to secondary evidence and force them to be assimilated into the probabilistic patterns of human language, which inevitably creates a semantic bottleneck.
In contrast, we argue that rather than serving as an external plugin, the impressive path forward lies in endogenous integration, unifying GFM and LLM into a cohesive whole.
By inherently anchoring linguistic tokens to a topological manifold, the generative process becomes strictly bounded by structural dependencies grounded in the cognitive substrate of GFM. Such integration would empower next-generation large models to achieve enhanced capabilities for complex logic and reasoning.

\section{Conclusion}
The road to Graph Foundation Models is currently divided, with both prevailing approaches facing inherent challenges. This paper illuminates a new pathway---Riemannian geometry as the foundational principle for GFM. It is distinguished from current Riemannian approaches that mostly focus on the exogenous representation space orthogonal to cross-domain graph generalities. We introduce the blue sky idea of RFM featured by intrinsic geometry  and endogenous structural inference. We showcase a progressive research agenda, redefining what a foundation model for graphs can be. Rather than incremental evolvement, RFM heralds a transformative paradigm shift that leads the graph domain into its large model era.


\bibliographystyle{ACM-Reference-Format}
\bibliography{KDD26BlueSky/main}










\end{document}